\documentclass{article}

\usepackage{arxiv}
\usepackage[utf8]{inputenc} % allow utf-8 input
\usepackage[T1]{fontenc}    % use 8-bit T1 fonts
\usepackage{hyperref}       % hyperlinks
\usepackage{url}            % simple URL typesetting
\usepackage{booktabs}       % professional-quality tables
\usepackage{amsfonts}       % blackboard math symbols
\usepackage{nicefrac}       % compact symbols for 1/2, etc.
\usepackage{microtype}      % microtypography
\usepackage{lipsum}
\usepackage{graphicx}
\usepackage{mathtools}
\usepackage{amsmath}

\graphicspath{ {./images/} }

\title{Multi-label Relation Modeling in Facial Action Units Detection}

\author{
Xianpeng Ji, Yu Ding, Lincheng Li, Yu Chen, Changjie Fan \\
  Fuxi AI Lab \\
  Netease\\
  Hangzhou, China \\
  \texttt{\{jixianpeng, dingyu01, lilincheng, chenyu5, fanchangjie\}@corp.netease.com} \\
}

\begin{document}
\maketitle
\begin{abstract}
%Facial Action Units Detection (FAUD), based on the Facial Action Coding System (FACS), makes the subtlety of human emotions available in various applications, such as micro-expression recognition and expression generation. Therefore, FAUD has recently become a popular research field. Inspired by the recent advance in text multi-label classification task, we adapt the sequence-to-sequence method for multi-label text classification, which directly models the relationship between labels to treat the multiple activated AUs as a sequence in the context of data representation, thus transforming the multi-label classification task into a sequence modeling task. We implement the above algorithm on the data set released by the FG-2020 Competition: Affective Behavior Analysis In-the-Wild (ABAW).
%%Facial Action Units Detection (FAUD), based on the Facial Action Coding System (FACS), makes the subtlety of human emotions available in micro-expression recognition and expression generation.
%%Inspired by the recent advance in text multi-label classification task, we adapt the sequence-to-sequence method for multi-label text classification, which directly models the relationship between labels to treat the multiple activated AUs as a sequence in the context of data representation, thus transforming the multi-label classification task into a sequence modeling task. We implement the above algorithm on the data set released by the FG-2020 Competition: Affective Behavior Analysis In-the-Wild (ABAW).
This paper describes an approach to the facial action units detection. The involved action units (AU) include AU1 (Inner Brow Raiser), AU2 (Outer Brow Raiser), AU4 (Brow Lowerer), AU6 (Cheek Raiser), AU12 (Lip Corner Puller), AU15 (Lip Corner Depressor), AU20 (Lip Stretcher), and AU25 (Lip Part). Our work relies on the dataset released by the FG-2020 Competition: Affective Behavior Analysis In-the-Wild (ABAW). The proposed method consists of the data preprocessing, the feature extraction and the AU classification. The data preprocessing includes the detection of face texture and landmarks. The texture static and landmark dynamic features are extracted through neural networks and then fused as the feature latent representation. Finally, the fused feature is taken as the initial hidden state of a recurrent neural network with a trainable lookup AU table. The output of the RNN is the results of AU classification. The detected accuracy is evaluated with 0.5$\times$accuracy + 0.5$\times$F1. Our method achieves 0.56 with the validation data that is specified by the organization committee. 
\end{abstract}
% keywords can be removed
\section{Introduction}
The task of Facial Action Units Detection (FAUD) aims at recognizing the subtle facial action units (AUs) defined in Facial Action Coding System (FACS)\cite{article}. It has been widely utilized in micro-expression recognition and expression generation. Recently, FAUD has become a popular research field\cite{kollias2017recognition,kollias2020analysing,zafeiriou2017aff,kollias2019expression,kollias2018multi,kollias2019deep}.
  
This paper describes a novel approach to FAUD. The proposed method consists of the data preprocessing, the feature extraction, the and AU classification. The data preprocessing relies on face segmentation and facial landmark detection. The face segmentation is to crop face region, retaining face expression but leaving out the useless background region. The facial landmark detection is to specify the predefined key facial points that are consistent with a specific facial expression. Two types of facial features are employed in our work. The one is the static features that reflect the texture information of a facial expression in a still image. The extraction of the static features is carried out from the segmented face. The other one is the dynamic features that reflect the movement (e.g. cheek raising) during a few successive time frames. The dynamic features are refined from the detected facial landmarks. In the AU classification, the static and dynamic features are fused and then taken to judge whether the target AUs are observed.
 
The current work focuses on the detection of AU1 (Inner Brow Raiser), AU2 (Outer Brow Raiser), AU4 (Brow Lowerer), AU6 (Cheek Raiser), AU12 (Lip Corner Puller), AU15 (Lip Corner Depressor), AU20 (Lip Stretcher), and AU25 (Lip Part)\cite{article}. AU1, AU2 and AU4 describe the eyebrow movements and AU15, AU20, and AU25 depict the mouth motions. A facial expression is often described with the co-occurrence of several AUs. For example, happy and surprise may be expressed with the co-occurrence of AU1 and AU2. In our work, the AU co-occurrence is modeled as a problem of sequence processing in the AU classification.
\section{Proposed Method}
\label{sec:headings}
\begin{figure} % picture
    \centering
    \includegraphics[width=6.2in]{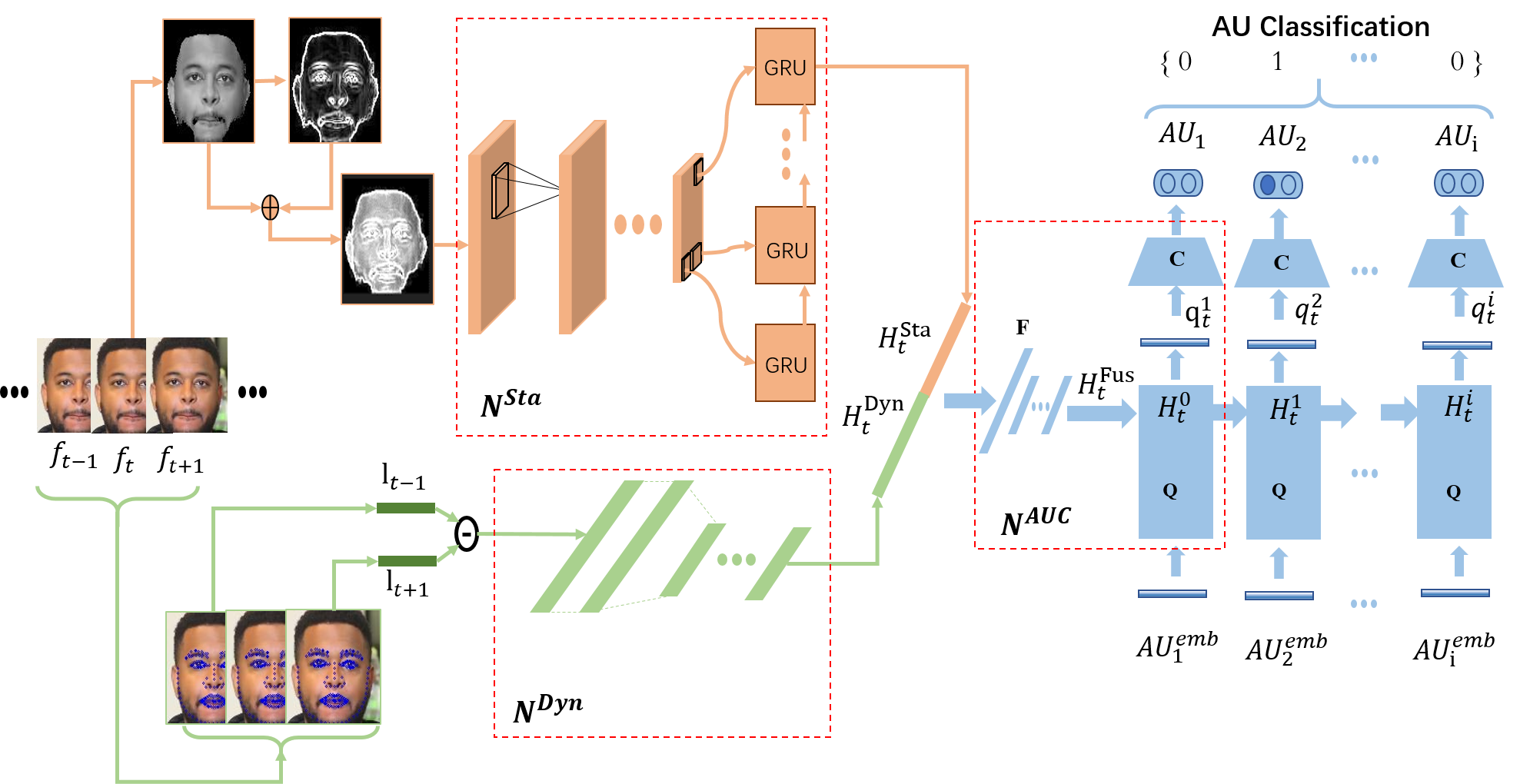}
    \caption{Pipeline of the proposed AU classification.}
    \label{fig:fig1}
\end{figure}

Our task is to detect the mentioned-above AUs for each image in a face video. Our AU classifier performs through the data processing, the static and dynamic features extraction and the AU classification. Particularly, an end-to-end deep neural network is proposed to carry out the static and dynamic features extraction and the AU classification. Its architecture is illustrated in Figure \ref{fig:fig1} and it will be detailed in this section.

\subsection{Data processing}
%%The AU process consists of three phases: onset, maintaining, and offset. For the onset and offset phases, An AU could be recognized by the temporal change of the specific area on the face, which could be effectively represented by the facial landmarks\cite{kollias2019deep}. Moreover, landmarks can also decouple the head pose factor. Hence, as is shown in Figure\ref{fig:fig1},  we employ a facial landmark detector on the results of face detection and tracking. The difference between frame $f_{t-1}$ and frame $f_{t+1}$ in landmarks is conducted to represented the dynamic feature of the target frame $f_{t}$ .
%% For the maintaining phase, although there is no obvious movement of face attribute, AU-related face texture could be obviously detected, namely the static feature.
%%Given the fact that there are still much backgrounds in the result of face detection and tracking, We further use a pre-trained face semantic segmentation model to remove the background. The segmented image is converted to grayscale as AUs have little relation to color information, then we use Sobel operator to extract the edge information from the image as another feature term. Based on the similarity in structure and the correlation in space, the segmented face and edge information are concatenated in channel dimension as the integrated feature.
Given a still image, a pre-trained face semantic segmentation model\footnote{https://github.com/Rijul-Gupta/face-seg} is employed to pick up the face texture region and mask the remaining background region with black. Then, the face texture is converted to a grayscale image and an edge image. A facial landmark detector is utilized to detect 73 face landmarks in a facial expression. The grayscale image and the edge image ignore part of face texture information but highlights a face texture structure of a still facial expression. A sequence of landmarks is dedicated to reflecting a face motion.
\subsection{Static Feature Extraction Network}
%%The overall structure of the proposed model is shown in Figure \ref{fig:fig1}. A cnn-based module S-net is responsible for extracting the static feature of appearance. Correspondingly, D-net outputs the dynamic feature based on the landmarks.Then, FQP-net  predicts the activation of each AU. 
A static network, denoted as $N^{Sta}$, is proposed to refine the static features from the grayscale image and the edge image. $N^{Sta}$ consists of a convolutional neural network (CNN) and a Gate Recurrent Unit (GRU). Particularly, CNN adopts limited receptive fields to specifically capture the local correlation between AUs and the local face regions. The grayscale image and the edge image are viewed as two input channels to CNN. Furthermore, to take into consideration the global correlation between AUs and the full face texture, the GRU is used to sequentially model the flatten feature map from the local correlation output of CNN\cite{Niu2019LocalRL}. The final hidden state, $H_t^{Sta}$, of the GRU is viewed as the output static features. The static feature extraction is formatted as follows. 
  
\begin{equation}
H_t^{Sta} = GRU(Covs(I_t,{\theta _c}),{\theta _g}) 
\end{equation}
  
%%where $Covs$ represents the convolution in the CNN with the parameter $\theta_c$ to be learned, and $I_t$ is the segmented face and edge information of current frame cascaded in channel dimension. The module based on GRU with parameter $\theta_g$ outputs $H_t^c$. 
where $Covs$ represents the operation of convolution in the CNN with the parameters $\theta_c$, and $I_t$ represent the input of grayscale image and edge image in two channels. $H_t^{Sta}$ is the output static features from the GRU with parameters $\theta_g$. 
%%\subsubsection{CNN-based Static Appearance Module S-net}
\subsubsection{Dynamic Feature Extraction Network}
%%Extracting and combining the features of specific facial areas for corresponding AUs are thought significant to de-noise and enhance the discrimination for FAUD\cite{kollias2020analysing}.
%%Generally, local correlation between AU and facial region, as prior-knowledge, is implemented in a patch-learn manner. For simplification, the convolutional neural network (CNN) with limited receptive field is adopted. Moreover,inspired by\cite{Niu2019LocalRL}, in order to maintain spatial structure information of the full image, a local relationship learning module based on Gate Recurrent Unit (GRU) is introduced to serially take in the flattened feature map from CNN by regions, after which the last hidden state  $H_n$ of GRU could be seen as a representation of the fully image with the regularization of face spatial structure.
A dynamic network, denoted as $N^{Dyn}$, is proposed to extract the dynamic features at time frame $t$ from the difference between the landmarks of two neighboring images at time frames $t$-$1$ and $t$+$1$.

A full connected neural network, $f$, takes the landmark difference as input and outputs the dynamic features. The last layer of $f$ is a non-linear activation of $Tanh$ to make the dynamic features with the scale as same as the static features. The dynamic features extraction is formatted as follows.
  
\begin{equation}
H_t^{Dyn} = f({diff}_{t},{\theta _D})
\end{equation}
  
where $\theta_D$ refers to the parameters of $N^{Dyn}$ and the $diff_{t}$ stands for the landmark difference at the time frame $t$.

%%\begin{equation}
%%H_t^c = GRU(Covs(I_t,{\theta _c}),{\theta _g}),
%%\end{equation}
%%where $Covs$ represents the convolution in the CNN with the parameter $\theta_c$ to be learned, and $I_t$ is the segmented face and edge information of current frame cascaded in channel dimension. The module based on GRU with parameter $\theta_g$ outputs $H_t^c$. 
%\subsubsection{ Landmark-based Dynamic Shape Module D-net}
\subsubsection{AU Classification Network}
%%The difference between the 73 landmarks of the frame around the current frame could represent its dynamic feature. To cope with the shake of the face detection and tracking, we employ a two-layer linear transformation aiming to learn an affine transformation according to the landmarks not affected by the AUs.
%%\begin{equation}
%%diff_t = l_{t + s} - l_{t - s}, t \in (s,n - s),
%%\end{equation}
%%where  $l$  , as a vector, is the landmarks coordinates of the corresponding frame.  $s$ is the window size sliding on frames of videos, and could be treated as a hyper-parameter, set to be 1 in this work. $n$ is the number of frames of a certain video. 
%%To get the feature about AUs in form of GRU hidden state, a fully-connected neural network $f$ with a non-linear activation of $Tanh$ is employed, formulated as 
%%\begin{equation}
%%H_t^l = {\mathop{\rm Tanh}\nolimits} (f(diff_t,{\theta _D})),
%%\end{equation}
%%where $\theta_D$ is the parameters of D-net.
An AU classification network, denoted as $N^{AUC}$, is designed to judge the occurrence of AUs. It is decomposed of three sub-networks including the feature fusion network $F$, the query-based network $Q$ and the classification network $C$. 

The feature fusion network is to fuse the static features and the dynamic features. Its output, $H_t^{Fus}$, is a representation of the fused static and dynamic features.
The query-based network consists of a learnable AU embedding query table and a GRU. The AU embedding table is 8$\times$64 matrix. Each column refers to an AU embedding. The size of the matrix refers to the number of the considered AUs and the dimension of an AU embedding. $GRU$ takes $H_t^{Fus}$ as its initial hidden state in the AU classification, $H_{t}^{0}$. When queried by the embeddings of AUs in a fixed order, $GRU$ outputs the result to a binary classifier $C$ and recurrently update $H_{t}^{i}$. The output of GRU based on each $H_{t}^{i}$ is used to judge whether an AU occurs. $C$ is a fully-connected neural network and it is to judge the activation of the AUs rely on the current query of a specific AU embedding. The AUs classification could be formatted as follow.
  
\begin{equation}
\begin{aligned}
q_t^i,H_t^{i + 1} &= Q(AU_i^{emb},H_t^i,{\theta _q})\\
H_t^{Fus} &= H_t^{Dyn} \cup H_t^{Sta}\\
p_t^i &= softmax (C(q_t^i,{\theta _p}))
\end{aligned}
\end{equation}
  
where  $i$ is the index of AUs. $\cup$ represents the fusion of the static and the dynamic features. $AU_i^{emb}$ is the embedding of $i$-$th$ AU. $\theta _q$ is the parameters of $Q$ and $\theta _p$ is the parameters of $C$. they are learned jointly.

\section{Results}
\label{sec:others}

%%Based on the cropped and aligned dataset provided by the competition, we get all the pre-processing intermediate features of the training data according to the above pre-processing method. The dataset is reconstructed based on the video frames. When building the model, we set the dimension of AU embedding to 64, same to the dimension of every GRU-based network mention above. The batch size is set to 128. In the process of a end-to-end training, random sampling on the reconstructed training data is carried out.
%%\subsection{Experiments result}
%%We employ early stopping after convergence. According to the evaluation method of 0.5 *accuracy + 0.5 * F1, the performance of the model on the validation dataset is 0.53 without post-processing and hyperparameter tuning.
According to the evaluation metrics of 0.5$\times$accuracy + 0.5$\times$F1, the performance on the validation dataset of the dataset Aff-Wild2\cite{kollias2018aff} is 0.56 with hyper-parameters tuning and post-processing of smoothing. More details about our experiments will be reported in the near future version.

\bibliographystyle{unsrt}  
\bibliography{references.bib}
\end{document}